\documentclass[journal]{IEEEtran}

\ifCLASSINFOpdf
\else
\fi
\usepackage{amsmath}
\usepackage{booktabs}
\usepackage{caption}
\usepackage{subcaption}
\usepackage{todonotes}
\usepackage{adjustbox}
\usepackage{makecell}
\usepackage{graphicx}
\usepackage{multirow}
\usepackage{algorithm,algorithmic}
\usepackage{xspace}
\usepackage{makecell}
\usepackage{float}
\usepackage{placeins}
\usepackage{enumitem}

\begin{document}
%
\title{Human Comfortability Index Estimation in Industrial Human-Robot Collaboration Task}


\author{Celal~Savur,~\IEEEmembership{Member,~IEEE,}
        Jamison~Heard,~\IEEEmembership{Member,~IEEE,}
        and~Ferat~Sahin,~\IEEEmembership{Senior,~IEEE}
        
\thanks{C. Savur is with Intel Labs, Hillsboro, OR, 97124 USA, {celal.savur@intel.com}}
\thanks{J. Heard and F. Sahin are with the Department
of Electrical and Microelectronic Engineering, Rochester Institute of Technology, Rochester,
NY, 14623 USA, {\{jrheee, feseee}\}@rit.edu}
\thanks{}
}

\maketitle

\begin{abstract}
Fluent human-robot collaboration requires a robot teammate to understand, learn, and adapt to the human's psycho-physiological state. Such collaborations require a computing system that monitors human physiological signals during human-robot collaboration (HRC) to quantitatively estimate a human's level of comfort, which we have termed in this research as \textit{comfortability index} (CI) and \textit{uncomfortability index} (unCI). Subjective metrics (\textit{surprise}, \textit{anxiety}, \textit{boredom}, \textit{calmness}, and \textit{comfortability}) and physiological signals were collected during a human-robot collaboration experiment that varied robot behavior. The emotion circumplex model is adapted to calculate the CI from the participant's quantitative data as well as physiological data. To estimate CI/unCI from physiological signals, time features were extracted from \textit{electrocardiogram} (ECG), \textit{galvanic skin response} (GSR), and \textit{pupillometry} signals. In this research, we successfully adapt the circumplex model to find the location (axis) of `comfortability' and `uncomfortability' on the circumplex model, and its location match with the closest emotions on the circumplex model. Finally, the study showed that the proposed approach can estimate human comfortability/uncomfortability from physiological signals.


\end{abstract}

\begin{IEEEkeywords}
Comfortability Index, Physiological Computing, Human-Robotic Collaboration, Circumplex Model
\end{IEEEkeywords}

%

\IEEEpeerreviewmaketitle

\section{Introduction}
The collaborative robot market is projected to see a substantial Compound Annual Growth Rate (CAGR) of approximately 57.7\% over the next five years \cite{Absolutereports2019}. As more of these robots are incorporated into the manufacturing environment, their unique capability to work alongside humans pushes the necessity of safety and efficiency to the forefront. One vital safety consideration pertains to potential injuries resulting from human-robot collisions.

\begin{figure}[h!]
    \centering
    \includegraphics[width=0.6\linewidth]{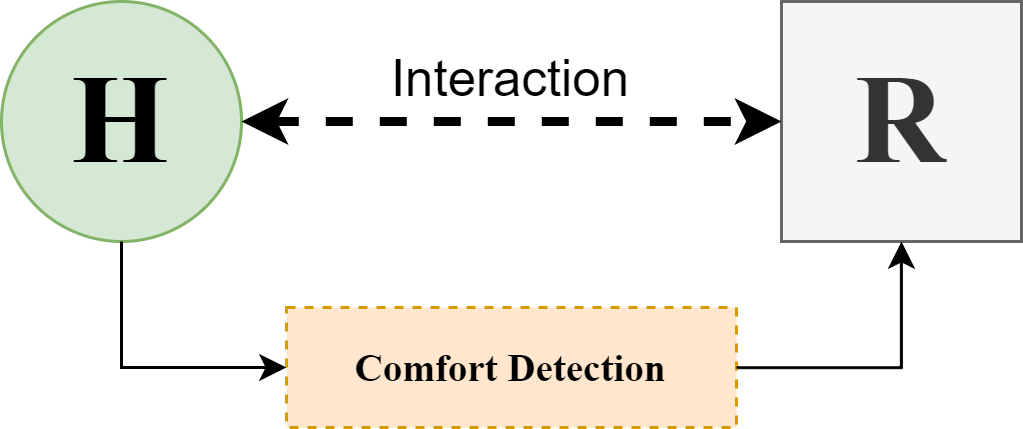}
    \caption{Proposed physiological computing system where ``R" represents a robot and ``H" represents a human.}
    \label{fig:proposed_system_overview}
\end{figure}

Numerous strategies exist to mitigate such collisions, ranging from simple solutions like physical and electronic safeguards to more intricate protection schemes like dynamic speed and separation monitoring \cite{ISO2016, Kumar2021}. While these methods effectively decrease injury risk, any physical collision or unexpected robot behavior during the collaborative process can impact the comfort level of the human operator. These impacts to humans should, as much as possible, be mitigated, in order to ensure the efficiency and comfort level of HRC tasks. To address this need, we propose a system capable of detecting human physiological signals in order to estimate their mental and physiological state \cite{NSF2019}. Such a system, known as a physiological or affective computing system \cite{Fairclough2017}, is illustrated in Fig. \ref{fig:proposed_system_overview}. Once this system estimates the human's mental and physiological state, these estimates can be integrated into personalized safety measures and adaptive human-robot collaboration strategies.

Few studies have explored the matter of human comfort during collaborative tasks with robots \cite{Kulic2007, Kulic2007a}. In our study, we investigate this matter by modulating the behavior of a robot in surprising and expected ways, and then analyzing the physiological and affective reactions of their human collaborators. In more precise terms, we employed a collaborative robot (cobot) in a human-robot cooperation task and in an experimental setting manipulated the robot's \textit{velocity}, \textit{trajectory}, and \textit{sensitivity}. Our goal was to gain a deeper understanding of how the robot's movements impacted human comfort under varying conditions.

The underlying premise is that a robot's behavior can influence the human's level of comfort during human-robot collaboration, and that human comfort level is an important factor in human-robot collaboration outcomes. For instance, a human may feel less comfortable when the robot's velocity is high as opposed to when it is slower.


The contributions of this research are three-fold:
\begin{itemize}
    \item Two novel approaches (circumplex model) and Kernel Density Estimation (KDE) model for estimation of the comfortability (CI) and uncomfortability index (unCI) was presented and their performances were evaluated.
    \item A physiological computing system that estimates the human CI/unCI from physiological signals was presented.
    \item Multiple Machine Learning (ML) algorithms were developed, and their performances were evaluated. 
 \end{itemize}

The paper is organized as follows. Physiological computing and related works are presented in Section \ref{sec:related_works}. Methodology, data collection, and data processing are discussed in Section \ref{sec:methodology}. The experiments and data collection devices are explained in Section \ref{sec:experiments}. The results and discussion for the proposed method are presented in Section \ref{sec:results}. Section \ref{sec:limitations} talks about limitations of the proposed research. Finally, the conclusion and future works are provided in Section \ref{sec:conclusion}.

\section{Related Works}
\label{sec:related_works}
Hu et al. devised a system that leverages EEG and GSR signals to estimate a real-time human trust index. The subjects in their experiment assessed a virtual sensor reading within a simulation \cite{Hu2016}. With sensor accuracy and the participant's reaction as key determinants, the researchers demonstrated the potential of physiological signals as a promising tool for gauging human trust levels.

Kulic et al. \cite{Kulic2005} examined anxiety induced by two different trajectory planners— a safe planner and a classical planner— implemented by an arm robot. The empirical evidence gathered from participants' responses revealed that the safe planner evoked lesser anxiety compared to the classical one \cite{Kulic2005a}. Building upon their previous work, Kulic and Croft \cite{Kulic2006, Kulic2007} demonstrated the superiority of the Hidden Markov Model (HMM) over Fuzzy inference in terms of accurately estimating arousal and valence from physiological signals. In a slightly different approach, Rani et al. \cite{Rani2007} adopted fuzzy logic and a regression tree for anxiety detection. They utilized a variety of physiological signals in their experiment such as Electrocardiogram (ECG), electrodermal activity (EDA), electromyograms (EMG), and temperature signals to facilitate implicit communication between a human and a robot.

Villani et al. \cite{Villani2018a} presented a framework designed to assess human stress levels based on heart-rate variability (HRV). This was geared towards enhancing human-robot interaction within an industrial context. During the study, they program the robot to reduce it is speed by 50\% when the human is stressed. While this extension in task completion time impacted efficiency, it optimized the interaction process. Similarly, Dobbins et al. \cite{Dobbins2018} used GSR signals during a commute to estimate stress during daily activity. In addtion, Tan et al. \cite{Too2009} conducted two distinct experiments; the impact of robot motion speed and the effects of varying distances between humans and robots \cite{Too2009}.

Heard et al. \cite{Heard2019} proposed a multi-dimensional algorithm that estimates human's workload using ECG signals, posture data, and environmental noise level. Furthermore, an operator's mental workload was examined during a teleportation task via HRV measurements by Landi et al. \cite{Landi2018}.

As computer games become more popular, Rani et al. \cite{Rani2006} attempted to make computer games more engaging by estimating a gamer's affective state to alter the game difficulty in real-time. The results showed that their system improved player's performance and lowered player’s anxiety during gameplay \cite{Rani2006}. Liu et al. \cite{Liu2006} designed a system that used affective cues to improve human-robot interaction by using various physiological signals such as ECG, EDA, and EMG to control the game's difficulty. As a result, the participant reported a 71\% increase in satisfaction while the anxiety-based adaptation system was active \cite{Liu2006}. 

In addition to the machine learning approach, Amin et al. \cite{Amin2019a} used variation in skin conductance to model the autonomic nervous system by using multi-rate formulation. Similarly, Wickramasuriya et al. \cite{Wickramasuriya2018} estimated stress using the Marked Point Process to track sympathetic arousal from skin conductance in \cite{Wickramasuriya2020}.


In all the aforementioned methods and corresponding studies, physiological signals were consistently reliable indicators for estimating the stress level, trust level, and workload of humans using machine learning algorithms or statistical tools. In this study, we were inspired by the James-Lange theory of emotion which indicates that our physical state changes before we experience an emotion \cite{x}, and leverage Russell's circumplex model to estimate human comfort level \cite{Russell1980}. This study is motivated by the potential to bring these schools of thought together; in this study we bridge research in physiological stress indicators with research on human comfort level. Early research by Toisoul et al. \cite{Toisoul2021} showed that estimation of human emotion from arousal and valence is possible. Hence, we incorporate emotions—surprise, anxiety, boredom, and calmness—to estimate \textit{arousal} and \textit{valence} (AV). Then, the proposed methods are used to determine the position of the comfortability and uncomfortability within the AV domain to estimate Comfortability Index (CI)/UnComfortability Index (unCI).

\section{Methodology}
\label{sec:methodology}
The circumplex model allows each emotion to be represented by \textit{arousal} (y-axis) and \textit{valence} (x-axis). The \textit{arousal} value indicates how calming or exciting the emotion is, and the \textit{valence} value indicates how negative or positive the feeling is. Thus, any emotion can be represented as a function of arousal and valence \cite{Russell1980}. The circumplex model can be divided into four quarters. For example, the emotion ‘happy’ is in the first quarter of the graph where both \textit{valence} and \textit{arousal} are positive. On the other hand, ‘sad’ is in the 3rd quarter where \textit{arousal} and \textit{valence} are negative. The strength of emotion is measured by the distance from the origin \cite{Toisoul2021, Mollahosseini2017}, where the origin represents neutral.

\begin{figure}[h!]
    \centering
    \includegraphics[width=0.8\linewidth]{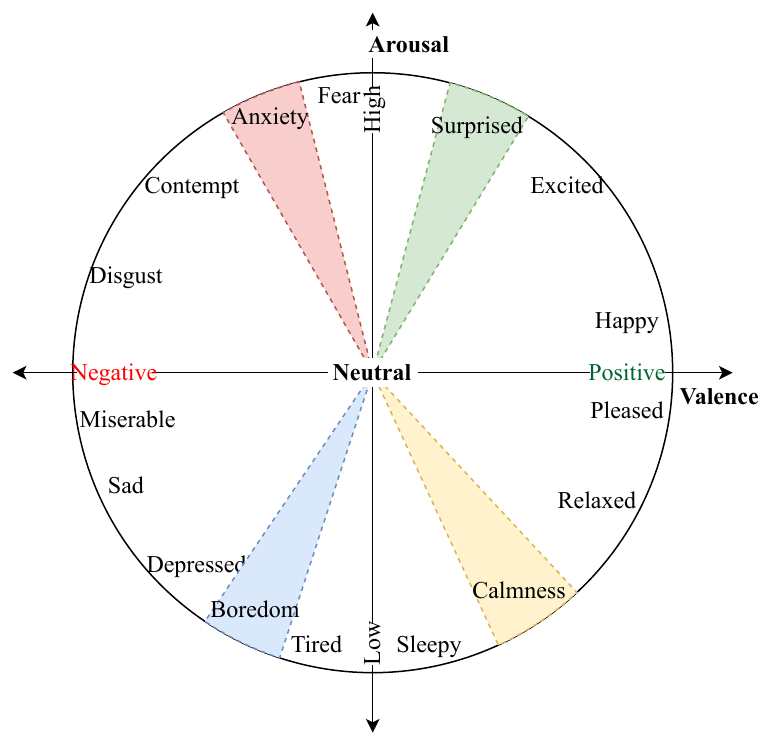}
    \caption{Arousal and Valence circumplex model and a few discrete emotional classes and their locations \cite{Russell1980, Toisoul2021}. }
    \label{fig:circumplex_emotion_loc}
\end{figure}

The basic emotions are marked in the circumplex model shown in Fig. \ref{fig:circumplex_emotion_loc}. As discussed in previous research, comfortability is different from an emotional or affective state, and it can be triggered by a combination of multiple emotions/affective states \cite{Elena2020}. Therefore, we used \textit{surprise}, \textit{anxiety}, \textit{boredom}, and \textit{calmness} emotions that locate in each quarter in the circumplex model (marked with color) to estimate comfortability and uncomfortability. 

\begin{figure}[h!]
    \centering
    \includegraphics[width=1.0\linewidth]{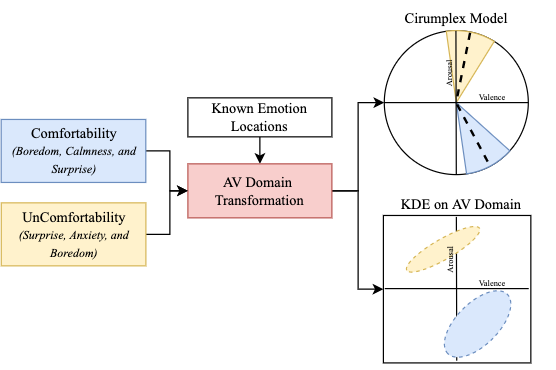}
    \caption{Two approaches: the first approach is a circumplex model and the second one is a Kernel Density Estimation (KDE) model.}
    \label{fig:proposed_aproaches}
\end{figure}

In this research, we focused on two approaches, Circumplex model and Kernel Density Estimation (KDE) model. The first approach assume the comfortability and uncomfortability are function of multiple emotions ant they can represented on circumplex representation (a.k.a emotion wheel). The second approach does not make any assumption about it only rely on the AV points to estimate distribution of comfortability and uncomfortability. 

The proposed methods shown in Fig. \ref{fig:proposed_aproaches} use reported emotions (\textit{surprise}, \textit{anxiety}, \textit{boredom}, and \textit{calmness}) and pre-defined emotion locations to estimate arousal and valence. Although both approaches use the same formula for the AV Transformation, the estimation of the comfortability index (CI) and uncomfortability index (UnCI) is different. The first approach estimates the comfortability and uncomfortability axis location (angles) on the circumplex model and the second approach uses AV data points to estimate the distribution of the data points from reported comfortability or uncomfortability values by the participants.

\subsection{AV Domain Transformation}
In this study, we are looking at comfortability and uncomfortability separately, because different emotions cause them. For example, when someone is feeling \textit{anxious}, they cannot be comfortable at the same time. But, interestingly, a person can feel \textit{surprised}, \textit{calm}, or even \textit{boredom} and still be comfortable.
In the same way, a person cannot be uncomfortable while also feeling \textit{calm}. But, a person might feel \textit{surprised}, \textit{boredom}, and \textit{anxious} but also could be uncomfortable. These perceived emotions were collected from participant via tablet (see section \ref{sec:data_collection}). 

In order to transform from emotions to AV domain, we assign the location of the \textit{surprise}, \textit{anxiety}, \textit{boredom}, and \textit{calm} emotions from other research \cite{Russell1980, Mollahosseini2017, Toisoul2021, Du2020} ($surprise=60^\circ$, $anxiety=110^\circ$, $boredom=240^\circ$, $calmness=290^\circ$) as shown in Fig. \ref{fig:circumplex_emotion_loc}. Then, $\pm 5$ degree noise was added to each emotion axis, as shown in Fig. \ref{fig:circumplex_emotion_loc}.

The $p_{e}$ and $\theta_{e}$ vectors use different emotions, in comfortability estimation, we used \textit{calmness}, \textit{surprise}, and \textit{boredom} emotions and \textit{surprise}, \textit{anxiety}, and \textit{boredom} emotions for uncomfortability estimation as shown in Fig. \ref{fig:proposed_aproaches}.

Afterward, the arousal and valence location $AV_{loc}$ is computed as
\begin{align}
    d_{valence} &= \frac{ \vec{p_{e}} \cdot cos(\vec{\theta_{e}})}{\sum \vec{p_{e}}} \label{eq:valence} \\
    d_{arousal} &= \frac{ \vec{p_{e}} \cdot sin(\vec{\theta_{e}})}{\sum \vec{p_{e}}} \label{eq:arousal}\\
    AV_{loc}    &= (d_{valence}, d_{arousal}) \label{eq:av_loc}
\end{align}

where the $d_{valence}$ indicates the \textit{valence} component of the $AV_{loc}$, and $d_{arousal}$ indicates the \textit{arousal} component of the $AV_{loc}$ for comfortability and uncomfortability separately. We devise each metric with the sum of emotions to normalize the estimated arousal and valence. 

\subsection{Circumplex Model}
\begin{figure}[h!]
    \centering
    \includegraphics[width=0.9\linewidth]{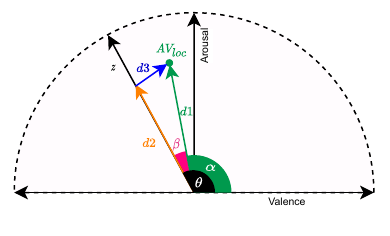}
    \caption{Given a $z$ axis on the circumplex model, how CI or unCI is calculated.}
    \label{fig:ci_un_ci_calculation}
\end{figure}

In the circumplex model, the origin represents a neutral/no feeling zone and emotions are getting stronger as they get close to the edge of the circumplex circle \cite{Toisoul2021, Mollahosseini2017}. Fig. \ref{fig:ci_un_ci_calculation} shows an $z$ axis and $AV_{loc}$. By following the circumplex approach and given an axis location and an $AV_{loc}$, we develop the formula \ref{eq:ci_un_ci_estimation} to estimate the strength of the CI/unCI by using $g(\theta, AV_{loc})$ as

\begin{align}
g(\theta, AV_{loc}) = \left\{
        \begin{array}{ll}
            d1 & \quad  if \quad \lvert \beta \rvert \leq 5^{\circ},  \\
            d2*(1-\frac{d3}{2})) & \quad  otherwise
        \end{array}
    \right.
    \label{eq:ci_un_ci_estimation}
\end{align}

where $\beta$ is an angle that calculated with absolute difference between $z$ axis' angle ($\theta$) and $AV_{loc}$'s angle ($\alpha$) as defined $\lvert \theta - \alpha \rvert $, $d1$ is the length of $AV_{loc}$ from the origin, $d2$ is the length of projection of $AV_{loc}$ on axis from the origin, and $d3$ is the distance between $AV_{loc}$ and the axis. 

In order to find the optimal $z$ axis that represents CI and unCI axis, we optimize $\theta$; the angle that minimizes the mean squared error (MSE) shown in \ref{eq:optimization}.
\begin{align}
minimize &= f_{0}(\theta) \nonumber\\
         &= \frac{1}{n}*\sum_{i=1}^{n}(y_{i} - g(\theta, a_{i})))^{2} \label{eq:optimization}
\end{align}
where $a_{i} \in R^{2}$ represents one $AV_{loc}$ that estimates in \ref{eq:av_loc}, $y_{i}$ is the corresponding comfort/uncomfort value that was reported by the participant.  

In summary, we use the AV locations to fit the best axis for comfortability and uncomfortability separately by using \ref{eq:optimization}. These two axes will be used to estimate the level of comfort and uncomfort.

\subsection{Kernel Density Estimation (KDE)}
The underlying assumption for the circumplex model was that an emotion is an axis where the emotion gets stronger as it moves to the edge of the circumplex circle. An alternative method can be to fit a distribution to the data points in the AV domain as shown in Fig. \ref{fig:proposed_aproaches}. Here, we used non-parametric kernel density estimation (KDE) to fit the data points in the AV domain by using the \textit{scipy.stats.gaussian\_kde} method with the \textit{weight} parameter set to the reported comfortability values \cite{Pedregosa2011}: 
\begin{align}
kernel      &= gaussian\_kde(AV_{loc}s, weight=w) \\
g(AV_{loc})	&= kernel(AV_{loc})/kernel_{max} \label{eq:kde_norm}
\end{align}
where $kernel$ is a fitted Gaussian KDE distribution, $w$ is the weight vector of perceived comfort or uncomfort levels, and $kernel_{max}$ is the maximum value of \textit{kernel}. We divide the result by the maximum value in \ref{eq:kde_norm} to have a comfortability/uncomfortability value that ranges [0, 1]. Fig. \ref{fig:kde} shows the KDE fitted to both the comfortability and uncomfortability levels. 


In summary, the KDE approach does not make any assumptions, it fits a non-parametric KDE distribution to the AV points. Then, we used the fitted kernel to estimate the likelihood to estimate the CI/unCI.

\subsection{Emotion Estimation from Physiological Signals}
\begin{figure}[h!]
    \centering
    \includegraphics[width=1.0\linewidth]{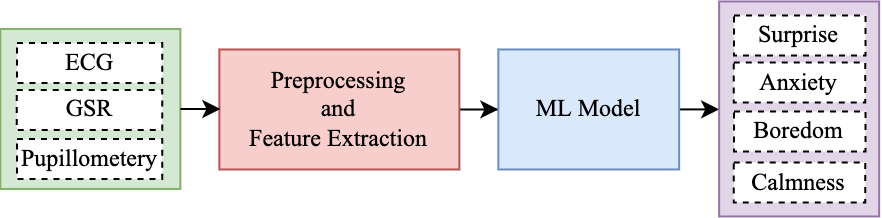}
    \caption{Emotion estimation from physiological signal.}
    \label{fig:emotion_estimation}
\end{figure}

In previous researches \cite{Hu2016, Kulic2005, Villani2018, Heard2019, Liu2006}, it has been shown that physiological signals such as EEG, ECG, GSR, and Pupillometry signals can be used in emotion detection, workload estimation, and anxiety detection. Hence in this study, we collected ECG, GSR, and Pupillometry. These signals were selected due to easy to collect and do not bother participants. The signals were collected in multiple trials during a human-subjects experiment. The signal length for each trial was 240 seconds approximately. These signals were used to estimate human emotions. As shown in Fig. \ref{fig:emotion_estimation}, the signals were pre-processed, and windowed, after which features were extracted. Then we applied two machine learning algorithms: Random Forest and Neural Network to predict emotions from physiological signals.

Brief information about the signals that were used is as follows:

\begin{itemize}[leftmargin=*]
    \item Electrocardiogram is a method to measure heart rate and heart rate variability (HRV). The ECG signal measures the electric activity of the heart by using electrodes that are placed near the heart. The important metrics that can be extracted from heart signal are heart rate and heart rate variability (HRV). Both metrics are calculated from consecutive peaks, which is called the RR interval for the ECG signal. 

    The time-domain metrics shown in Table \ref{tab:physiological_metrics} were used to quantify the ECG signal. Commonly used time-domain metrics are: mean of heart rate (mean HR), the mean of the RR (Mean RR), the standard deviation of the NN intervals (SDNN), the root mean square of the difference of consecutive NN intervals (RMSSD), and the proportion of a number of successive NN intervals that differ by more than 50 msec divided by the total number of NN intervals. 

    \item  The GSR (also known as skin conductance, electrodermal activity, or psychogalvanic reflex) is a method that measures skin conductance and provides information about the autonomic nervous system, which is responsible for preparing the body when we are stressed, nervous, surprised, or anxious \cite{Boucsein2012}. The GSR signal has two characteristics; tonic response and phasic response. The tonic response is slowly varying, and it depends on individual body features, such as skin dryness and autonomic regulation. The phasic response is sensitive to emotional and mental load \cite{Bradley1994a, Yik2011} and has a delay of 1--5 seconds after the onset of event stimuli. The GSR signal is usually used with other signals such as eye-tracking, heart rate, or respiration rate for better interpretation of the autonomic nervous system.

    A low pass filter with $order = 4$, and $Wn = 0.078$ was applied to the GSR signal, and a `boxzen' filter was used to smoothen the signal \cite{Carreiras}. Subsequently, the signal is decomposed into two components known as phasic (SCR) and tonic response. The phasic signal was used to detect onsets, peaks, peak amplitude, and recovery time. Gamboa's algorithm \cite{Gamboa2008} was used to detect onsets in the phasic signal. Common features extracted from the GSR signal are listed in Table \ref{tab:physiological_metrics}. These features are the mean and standard deviation of the tonic and phasic response, rate of onset event from the phasic response, mean of the peak amplitude, mean of the rise time, and mean of the recovery time.

    \item Pupil dilation is a measurement of pupil diameter changes, which can be affected by ambient light, other light intensity changes, emotional change, cognitive load, and arousal. It can also be used to detect emotional change \cite{Bonifacci2015, Mathot2018, Paprocki2017}. Pupillometry signals were recorded from both eyes. To remove high-frequency noise, a Butterworth low pass filter (cutoff=3, and order=2) was applied to the pupillometry signal. Then, time-domain features (see Table \ref{tab:physiological_metrics}) were extracted.

\end{itemize}

\begin{table}[h!]
\centering
\caption{Physiological metrics extracted from the ECG, GSR and Pupillometry signals}
\resizebox{\linewidth}{!}{\begin{tabular}{|c|l|l|}
\hline
\textbf{Type}                   & \textbf{Metric}    & \textbf{Description \& Unit}                                                                                            \\ \hline
\multirow{5}{*}{\textbf{ECG}}   & Mean HR            & Mean of Heart rate (bpm/min)                                                                                            \\ \cline{2-3} 
                                & Mean RR            & Mean of RR/IBI intervals (ms)                                                                                           \\ \cline{2-3} 
                                & SDNN               & Standard deviation of RR/IBI intervals (ms)                                                                             \\ \cline{2-3} 
                                & RMSSD              & \begin{tabular}[c]{@{}l@{}}The root mean square of the difference \\ of consecutive RR/IBI intervals (ms)\end{tabular}  \\ \cline{2-3} 
                                & pNN50              & \begin{tabular}[c]{@{}l@{}}Percentage of successive RR/IBI\\ intervals that differ by more than 50 ms (\%)\end{tabular} \\ \hline
\multirow{8}{*}{\textbf{GSR}}   & Tonic Mean         & \begin{tabular}[c]{@{}l@{}}Mean of tonic component of GSR signal \\ (Micro-siemens) \end{tabular}                                                                   \\ \cline{2-3} 
                                & Tonic Std.         & \begin{tabular}[c]{@{}l@{}}Standard deviation of tonic component\\ of GSR signal (Micro-siemens)\end{tabular}           \\ \cline{2-3} 
                                & Phasic Mean        & \begin{tabular}[c]{@{}l@{}}Mean of phasic component of\\ GSR signal (Micro-siemens)\end{tabular}                        \\ \cline{2-3} 
                                & Phasic Std.        & \begin{tabular}[c]{@{}l@{}}Standard deviation of tonic component\\ of GSR signal (Micro-siemens)\end{tabular}           \\ \cline{2-3} 
                                & Onset Rate         & SCR onset rate per second (onset/sec)                                                                                   \\ \cline{2-3} 
                                & Peak Amp. Mean     & Mean of Peak amplitude (SCR)(Micro-siemens)                                                                             \\ \cline{2-3} 
                                & Rise Time Mean     & Mean of rise time (SCR)(ms)                                                                                             \\ \cline{2-3} 
                                & Recovery Time Mean & Mean of recovery time (SCR)                                                                                             \\ \hline
\multirow{2}{*}{\textbf{Pupil}} & Pupil Mean         & Mean of pupil size                                                                                                      \\ \cline{2-3} 
                                & Pupil Std.         & Standard deviation of pupil size                                                                                        \\ \hline
\end{tabular}}
\label{tab:physiological_metrics}
\end{table}

\subsubsection{Random Forest Regressor}
The Random Forest (RF) regressor is a tree-based learner that uses multiple sub-samples of the dataset to train weak learners and uses the average of the weak learners for prediction \cite{Ethem2014}. An RF regressor was used for the estimation of \textit{surprise}, \textit{anxiety}, \textit{boredom}, and \textit{calmness}. To find the best parameters, grid-search with cross-validation was applied to the training dataset. The best parameters were selected as: the number of estimators (\textit{n\_estimator = 100}), criterion (\textit{mean squared error}), and the max depth (\textit{max\_depth = 6}). The rest of the configuration in scikit-learn was kept as default.

\subsubsection{Functional Neural Network}
\begin{figure}[h!]
    \centering
    \includegraphics[width=0.75\linewidth]{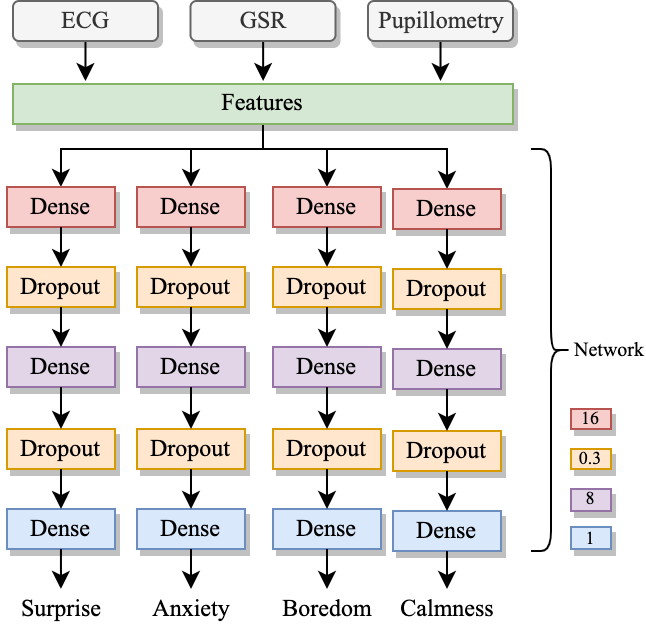}
    \caption{Functional Neural Network diagram that estimates emotions}
    \label{fig:nn_model}
\end{figure}

A functional Neural Network (NN) regression model that estimates \textit{surprise}, \textit{anxiety}, \textit{boredom}, and \textit{calmness} from physiological signal was developed. All extracted features in table \ref{tab:physiological_metrics} were normalized before use in training. Each branch of the NN model consists of three layers, which have 16, 8, and 1 neurons; respectively. To prevent over-fitting, two dropout layers (0.3) were added between the hidden layers in Fig. \ref{fig:nn_model}. The number of layers were selected by trial and error with the least complex model being selected. The \textit{Adam} optimizer with \textit{huber} loss was used for training. The data was randomly divided into training (17 subject's data) and testing (3 subject's data). The test set was never used during training. 

In summary, we estimated emotions from physiological signals using ML models, then we used equation \ref{eq:av_loc} to transform the emotion domain into the AV domain, after that equation \ref{eq:ci_un_ci_estimation} and \ref{eq:kde_norm} were used to estimate the CI/unCI level.

\section{Experiments}
\label{sec:experiments}
This experiment is a sequential collaboration since the participant waits for the robot to provide a part for the assembly \cite{BAUER2008}. The data were collected from healthy college students (N=20). The participants consisted of 17 male and 3 female subjects (\textit{Mean Age}= 24.70, \textit{SD}= 2.99) who has various experience with robotics. The experiment was approved by the Human Subject Research office at the Rochester Institute of Technology and the informed consent from each participant was collected before the experiment. 

\begin{figure}[h!]
	\centering
	\includegraphics[width=0.75\linewidth,keepaspectratio]{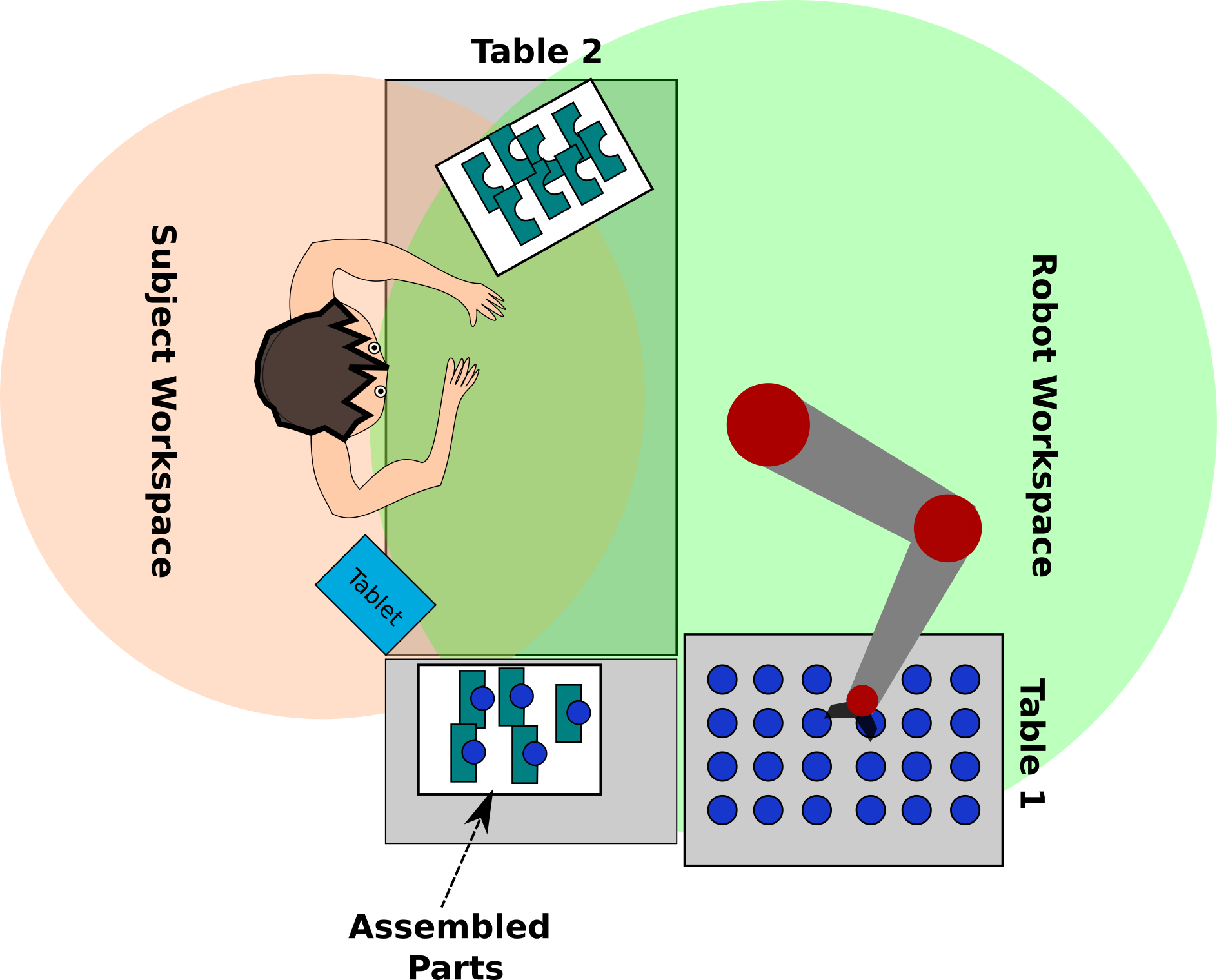}
	\caption{The experiment diagram is shown in a bird view.}
	\label{fig:exp_3}
\end{figure}

The experiment setup consisted of a joint task between the robot and the human. The task is an assembly task, where the robot provides a part from Table-1 and the human picks a part from Table-2 as shown in Fig. \ref{fig:exp_3}. The human is responsible for picking and screwing the two parts together, while the robot is holding one of the parts.

In the experiment, the robot changes its behavior by changing its \textit{velocity}, \textit{trajectory}, and \textit{sensitivity} (three independent variables). The velocity was set to be in two levels of \textit{normal} and \textit{fast}. These two modes were achieved by setting the global speed ratio to 0.7 for normal mode and to 1.0 for fast mode. The trajectory was defined as two modes: \textit{normal} and \textit{extreme}. The normal trajectory was defined as the robot moving from Table-1 to Table-2 with minimum joint movements. Hence, the normal trajectory was smooth and predictable. On the other hand, the extreme behavior passes multiple waypoints between pick and place locations, which makes the robot jerkier and less predictable. The sensitivity was defined as a threshold where robot will move the next steps based on the amount of force that participant apply to the end effector during the time the robot waits for the human to assemble the part. The sensitivity is defined in two modes: \textit{normal} and \textit{sensitive} and their threshold value set $\pm11 N/s^2$ and $\pm8 N/s^2$ respectively. 

This experiment consisted of 11 trials and one baseline. In a trial, the robot can provide up to 24 parts (iterations). The robot picks a part from \textit{Table 1} and moves it in front of the participant for assembling. The participant needs to pick an item from Table 2 and assemble the part in five seconds, then the robot moves to assembled part location drops the part, and moves to Table 1 for the next part. While the robot is dropping the part, the participant enters the subjective responses to the tablet where we asked the subject to report (touch screen) \textit{surprise}, \textit{anxiety}, \textit{boredom}, \textit{calmness}, and \textit{comfortability} level in a continuous value range [0, 1], as shown in Fig. \ref{fig:android_app}. The trial is completed when five minutes have elapsed or when there are no items left on Table 1 for the robot to pick up. The participant had an option to tap the robot's end effector to notify the robot not to wait anymore during assembling time. Hence, the participant can minimize the trial duration. Before the experiment started, all the participants were trained for one trial so that they get used to the task and the tablet. 

In this experiment, to have better signal labeling, a custom Android app was developed. The app allows a participant to enter their subjective responses immediately after each part assembly (iteration), the screenshot of the app can be seen in Fig. \ref{fig:android_app}. Hence, this approach produced more subjective data and a better idea of how the dependent variables are changing during the trials.

\subsection{Subjective Data Collection}
\label{sec:data_collection}
\begin{figure}[h!]
	\centering
	\includegraphics[width=\linewidth,keepaspectratio]{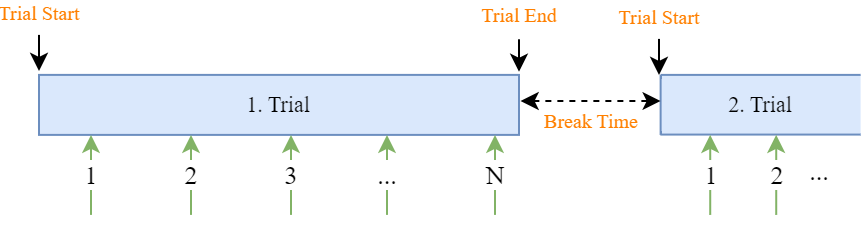}
	\caption{During trial data collection method. \cite{Sahin2022}.}
	\label{fig:data_collection_methods}
\end{figure}

\begin{figure}[h!]
	\centering
	\includegraphics[width=0.65\linewidth,keepaspectratio]{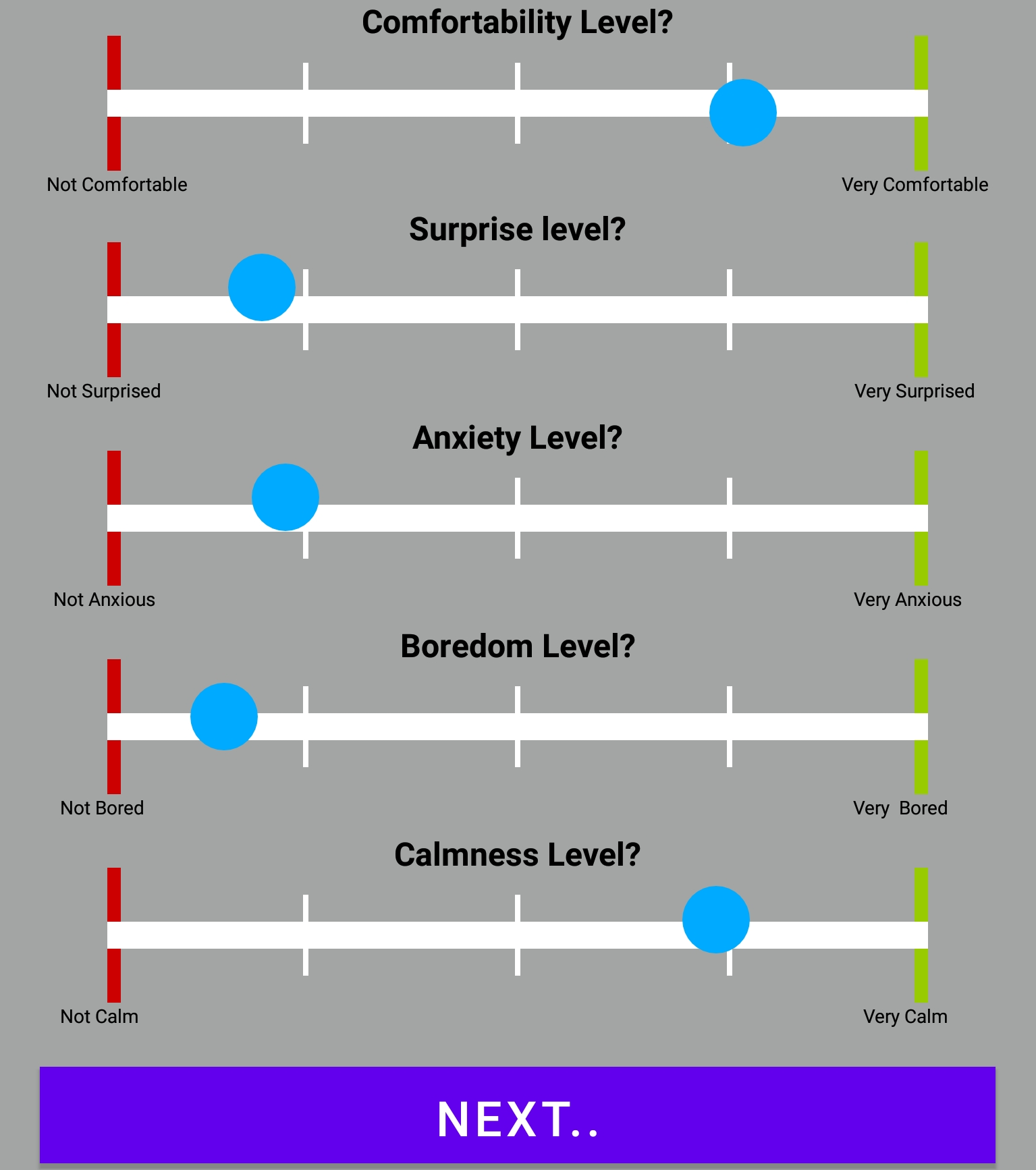}
	\caption{The screenshot of the custom Android app. The blue circle on the screen was randomly selected, it does not represent actual reporting. }
	\label{fig:android_app}
\end{figure}

In \textit{During-trial} data collection method participants reported their subjective responses multiple times throughout the whole trial \cite{Sahin2022} as shown in Fig. \ref{fig:data_collection_methods}. As this approach generated more labels and provides more information about how trial progress, it require time for the subject to report the subjective response. In order to minimize the reporting time, a custom Android app was developed and a screenshot can be seen in Fig. \ref{fig:android_app}. The app allows the participant to enter their subjective responses immediately after the assembly of each part (iteration) with a single tab on the tablet screen. This minimizes the duration of the collection, maintaining the integrity of the experiment.

\subsection{Data Labelling}
\label{sec:data_labelling}
We used \textit{during-trial} subjective responses as the label for the physiological signals. Since each iteration was taking 12 seconds on average we labeled six seconds before and after the subjective response. Hence, 12 seconds of signals were labeled according to the subjective response. This approach was applied to each trial, then we applied feature extraction to this 12 seconds window.

\subsection{Data Collection Devices}
The devices used for experiments were the Shimmer 3 GSR, Pupil Lab headset, and the Biopac BioHarness.
\begin{itemize}
    \item \textit{BioHarness} is a wireless chest strap that allows recording an ECG signal. In addition to the ECG, the device provides respiration rate, heart-rate, RR intervals, acceleration (3-axes), and device information.  
    \item \textit{Shimmer3 GSR+} is a widely used device in research due to its Bluetooth connectivity. The device provides one GSR channel that measures the conductance of the skin and one PPG channel \cite{Mejia-Mejia2020}. The sensor sampling rate was set to 128 Hertz (Hz). During the experiment, we asked participants to minimize the motion when they were using the hand that the sensor was placed.
    
    \item \textit{Pupil labs} headset is open-source hardware (eyeglasses) that has three cameras, two of which look at the eyes and one point to the subject's perspective \cite{Kassner2014}. The eye cameras operate at 120 frames per second (fps) and the world camera records at 30 fps. This device is widely used in research, and it provides a variety of signals such as pupil diameters, gaze location, and a real-time stream from the cameras.
    
\end{itemize}
Each sensor was calibrated for each subject independently. The subjects started with a baseline recording where the subject sits in front of the robot and was asked to relax for five minutes. The subjects who had vision correction were asked to remove them for better pupil signal quality. The pupil lab headset was connected to a Samsung S8 smartphone that has a Pupil Mobile app that transmits data to a local machine. The Shimmer3 GSR+ connects via Bluetooth to a computer, and a custom application \cite{Savur2019} developed to acquire these signals were used to send data using the Lab Stream Layer (LSL) protocol. The BioHarness was connected over Bluetooth to a computer as well. All devices were synchronized using the LSL stream library \cite{SCCN2018}. A modified version of the custom data collection app which generates automated event markers (\textit{trial start/stop}) and manual event markers was used \cite{Savur2019a}.

\begin{figure}[h!]
	\centering
	\includegraphics[width=\linewidth,keepaspectratio]{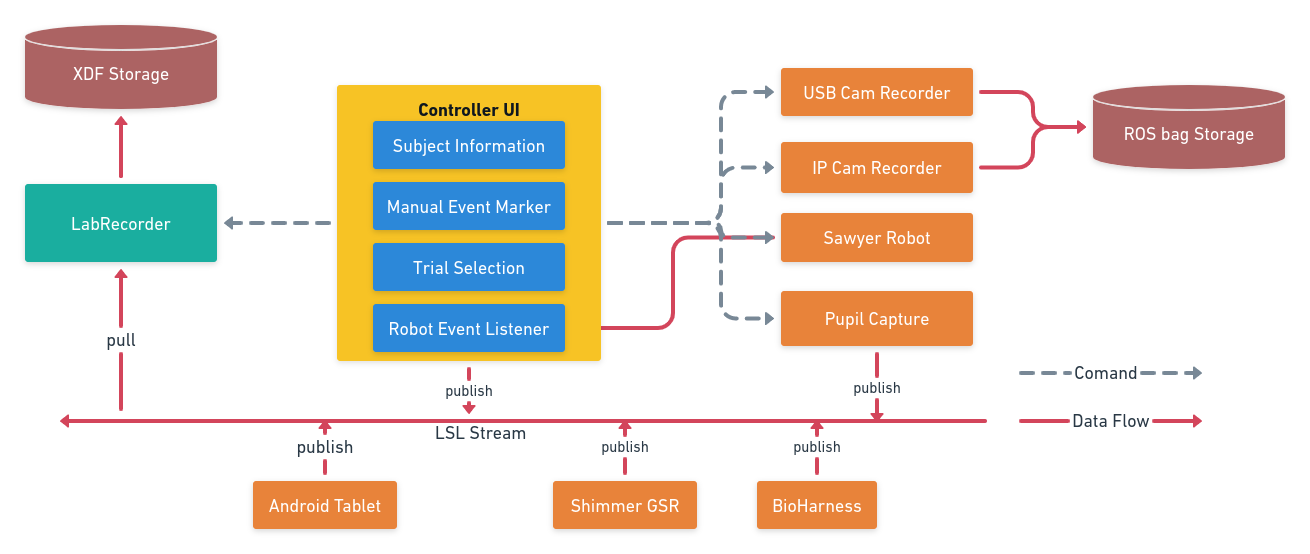}
	\caption{Overview of the Data collection system. The Controller UI is the main part of the system, which controls the experiment. It helps to gather subject information, generate event markers, launch all the other apps that record data, such as USB Cam recorder, Pupil Capture, etc. In addition, it toggles start/stop in LabRecorder to recorded LSL streams. All the data in the experiment was recorded in XDF files except camera streams and Pupil Camera streams. }
	\label{fig:data_flow}
\end{figure}

\subsection{Data Collection System}
\label{sec:data_collection}
Collecting signals from multiple devices and controlling the robot's behavior is a challenging task. There are multiple problems can arise such as synchronization, bandwidth, storage, and control. We extend our previous approach for data collection \cite{Savur2019} that can send commands to a robot to change its behavior on the fly. Fig. \ref{fig:data_flow} shows the command and data flow between multiple systems. The controller UI program is the main program that controls the entire data collection system. The controller is responsible to acquire the participant's information, trial selection, automated event generation, manual event generation, and send commands to the rest of the system. Lab Stream Layer (LSL) is a protocol that is widely used in time series data collection that has built-in time synchronization and distributed ability in the local network \cite{SCCN2018}. The LSL was used for signal transfer into the local network. The custom Android app also sends the subjective response to the LSL stream. The recording program (LabRecorder) has the ability to record all selected streams that are used for signal recording. For the video recording, we took another approach. The video from two streams USB cam and an IP cam recorded as a rosbag. The video data was used only for inspecting a trial if it was necessary.

\section{Results and Discussion}
\label{sec:results}
We evaluate the estimation of $CI_{axis}$ from the subjective responses, the calculation of $CI/UnCI$ using the circumplex model and KDE model, and the estimation of emotions from physiological signals by using multiple ML models. Further, the performance of the proposed approaches is tabulated and discussed.

\subsection{Comfortability Axis}
\begin{figure}[h!]
\centering
\begin{subfigure}{0.45\linewidth}
    \includegraphics[width=\linewidth]{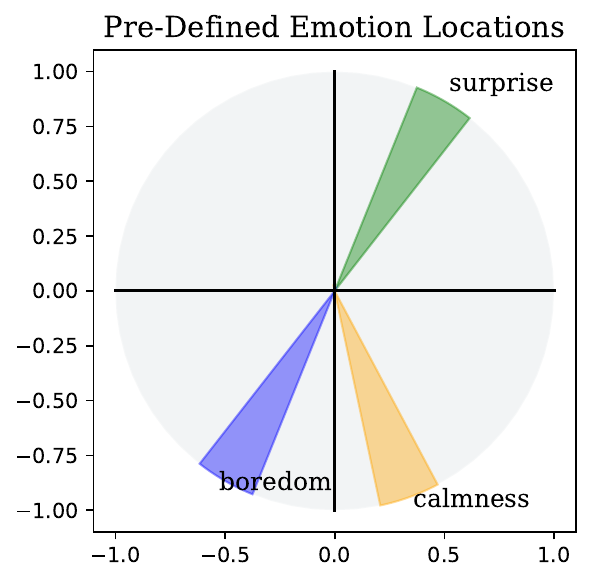}
    \caption{}
    \label{fig:comfort_emotion_loc}
\end{subfigure}
\begin{subfigure}{0.5\linewidth}
    \includegraphics[width=\linewidth]{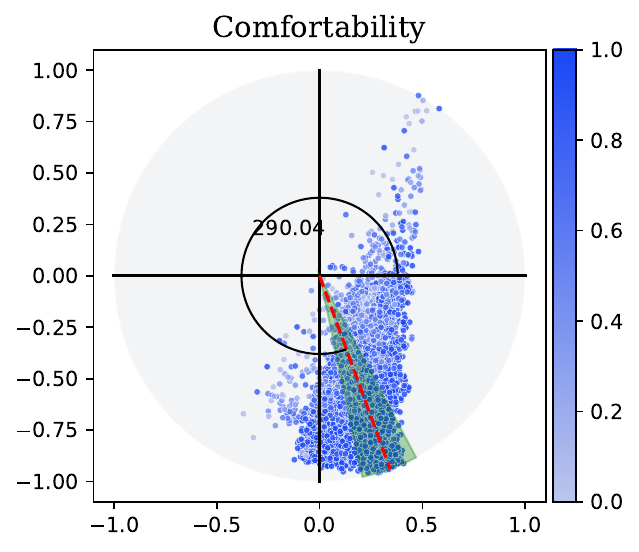}
    \caption{}
    \label{fig:comfort_axis_loc}
\end{subfigure}
\hfill
\caption{(a) shows the predefined emotion location for estimation of comfort axis, (b) indicates estimated CI axis location from subjective responses.}
\label{fig:figures}
\end{figure}

The subjective emotions (\textit{surprise}, \textit{boredom}, and \textit{calmness}) are shown in Fig. \ref{fig:comfort_emotion_loc} and were transformed into points (\textit{arousal} and \textit{valence} values) using equation \ref{eq:av_loc}. They were plotted in the AV domain as shown in Fig. \ref{fig:comfort_axis_loc}. Then, equation \ref{eq:optimization} was applied to the $AV_{loc}$s to estimate the best axis that represents the CI axis. As it can be seen in Fig. \ref{fig:comfort_axis_loc}, the CI axis is close to \textit{calmness}, that is expected since the subject reported the comfortability high when they are calm. 

\subsection{UnComfortability Axis}
\begin{figure}[h!]
\centering
\begin{subfigure}{0.45\linewidth}
    \includegraphics[width=\linewidth]{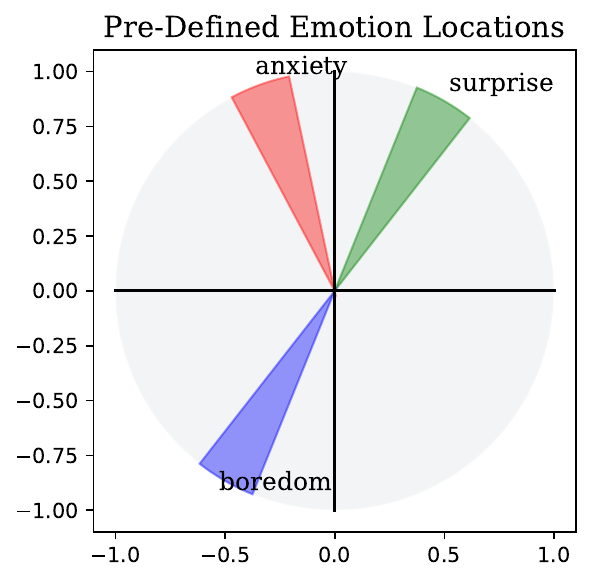}
    \caption{}
    \label{fig:uncomfort_emotion_loc}
\end{subfigure}
\begin{subfigure}{0.5\linewidth}
    \includegraphics[width=\linewidth]{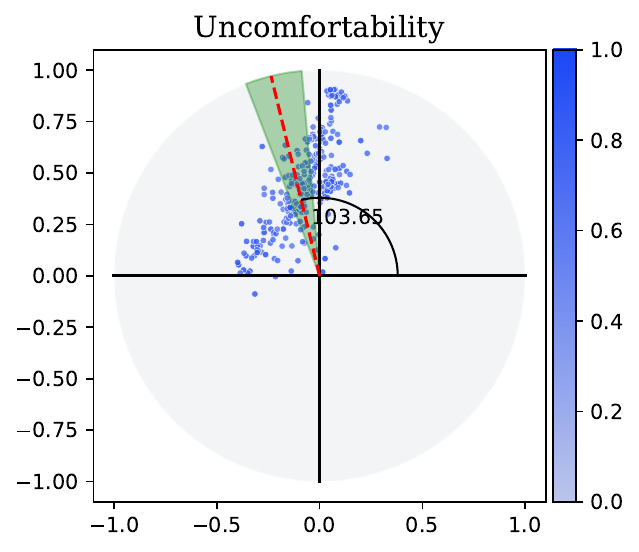}
    \caption{}
    \label{fig:uncomfort_axis_loc}
\end{subfigure}
\hfill
\caption{(a) shows the predefined emotion location for estimation of unCI axis, (b) indicates estimated unCI axis location from subjective responses.}
\label{fig:figures}
\end{figure}

Similar to the CI axis estimation, we used \textit{surprise}, \textit{anxiety}, and \textit{boredom} (see Fig. \ref{fig:uncomfort_emotion_loc}) emotions in \ref{eq:av_loc} to transform the emotions into the AV domain as shown in Fig. \ref{fig:uncomfort_axis_loc}. Then, equation \ref{eq:optimization} was applied to the $AV_{loc}$s to estimate the best axis that represent unCI axis. As it can be seen in Fig. \ref{fig:uncomfort_axis_loc}, the CI axis is close to \textit{anxiety}, that is expected since the subject reported high uncomfortability when they reported anxiety as high. 

\subsection{KDE Estimation}
\label{sec:kde_est}
\begin{figure}[h!]
\centering
\begin{subfigure}{0.48\linewidth}
    \includegraphics[width=\linewidth]{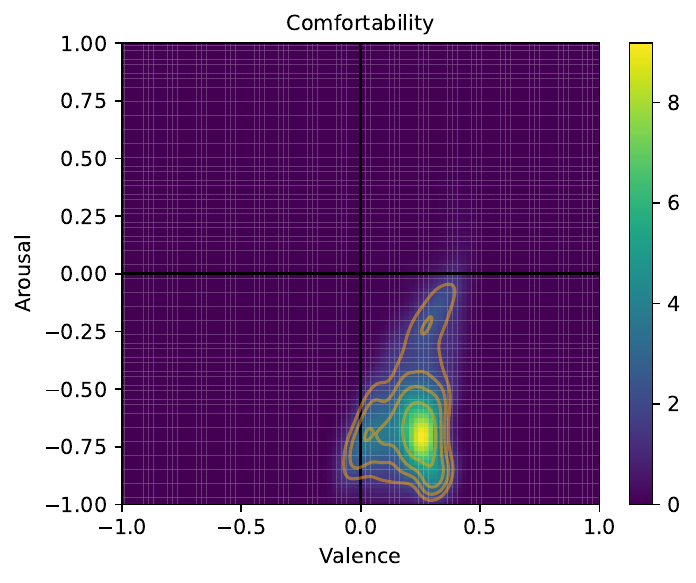}
    \caption{}
    \label{fig:comfort_kde}
\end{subfigure}
\begin{subfigure}{0.48\linewidth}
    \includegraphics[width=\linewidth]{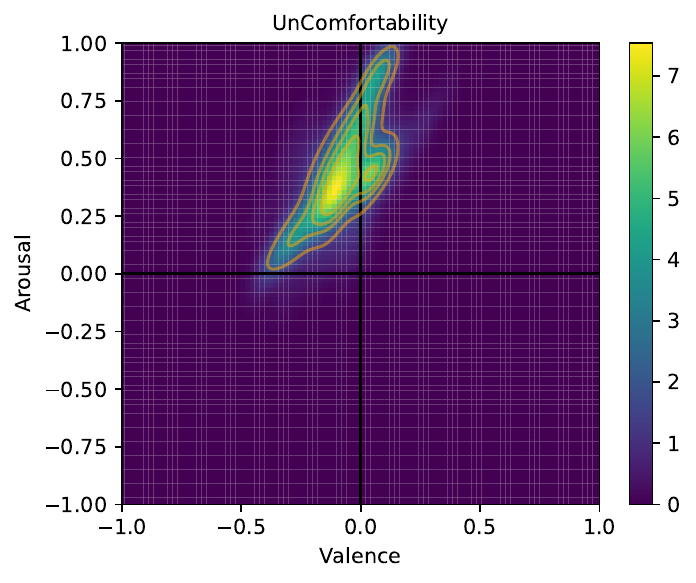}
    \caption{}
    \label{fig:uncomfort_kde}
\end{subfigure}
\hfill
\caption{(a) shows the KDE fitted to comfortability AV data points (b) shows the KDE fitted to uncomfortability AV data points.}
\label{fig:kde}
\end{figure}

Two KDE distributions were fitted to the perceived comfortability and uncomfortability AV data points. Fig. \ref{fig:kde} shows the fitted distribution in a heatmap. Unlike the circumplex model, the distribution for uncomfortability spreads into both quarters I and II. The comfortability distribution stays in quarter IV mostly.  

\subsection{Emotion and CI/unCI Estimation from Physiological Signals}
The dataset which consists of 2388 samples was used to develop the ML models. The dataset was divided into 70\% and 30\% for training and testing, respectively. Table \ref{tab:performance} shows the root-mean-square error (RMSE) and mean absolute error (MAE) of emotions, comfortability and uncomfortability for the Random Forest (RF) regressor, NN regressor, and Circumplex models. 

\begin{table*}[h!]
\centering
\caption{Performance of RF Regressor, NN, and Circumplex and KDE approaches.}
\label{tab:performance}
\resizebox{\linewidth}{!}{%
\begin{tabular}{|l|c|c|c|c|c|c|c|c|c|c|c|c|} 
\hline
\multicolumn{1}{|c|}{\multirow{2}{*}{\begin{tabular}[c]{@{}c@{}}\\\textbf{Models}\end{tabular}}} & \multicolumn{2}{c|}{\textbf{Surprise}} & \multicolumn{2}{c|}{\textbf{Anxiety}} & \multicolumn{2}{c|}{\textbf{Boredom}} & \multicolumn{2}{c|}{\textbf{Calmness}} & \multicolumn{2}{l|}{\textbf{Comfortability }} & \multicolumn{2}{l|}{\textbf{UnComfortability }}  \\ 
\cline{2-13}
\multicolumn{1}{|c|}{}                                                                           & \textbf{RMSE} & \textbf{MAE}           & \textbf{RMSE} & \textbf{MAE}          & \textbf{RMSE} & \textbf{MAE}          & \textbf{RMSE} & \textbf{MAE}           & \textbf{RMSE} & \textbf{MAE}                  & \textbf{RMSE} & \textbf{MAE}                     \\ 
\hline
RF                                                                                               & 0.20          & 0.15                   & 0.16          & 0.12                  & 0.14          & 0.11                  & 0.13          & 0.10                   & 0.16          & 0.11                          & 0.15          & 0.11                             \\ 
\hline
NN                                                                                               & 0.22          & 0.15                   & 0.18          & 0.12                  & 0.18          & 0.11                  & 0.15          & 0.10                   & 0.19          & 0.11                          & 0.19          & 0.11                             \\ 
\hline
Circumplex (RF)                                                                                  & -             & -                      & -             & -                     & -             & -                     & -             & -                      & 0.35          & 0.30                          & 0.28          & 0.24                             \\ 
\hline
Circumplex (NN)                                                                                  & -             & -                      & -             & -                     & -             & -                     & -             & -                      & 0.34          & 0.30                          & 0.26          & 0.24                             \\ 
\hline
KDE(RF)                                                                                          & -             & -                      & -             & -                     & -             & -                     & -             & -                      & 0.47          & 0.40                          & 0.51          & 0.43                             \\ 
\hline
KDE(NN)                                                                                          & -             & -                      & -             & -                     & -             & -                     & -             & -                      & 0.50          & 0.46                          & 0.60          & 0.54                             \\
\hline
\end{tabular}
}
\end{table*}

Table \ref{tab:performance} shows the RMSE and MAE of emotions, comfortability, and uncomfortability that are estimates from the RF, NN, and Circumplex model approaches by using physiological signals. The circumplex model is only used to estimate comfortability and uncomfortability. The circumplex model approach uses estimated emotion from both RF and NN. From the table, we can see that the RMSE and MAE are higher for the circumplex model than the RF and NN models. This is due to the estimation error that occurs when estimating emotion from two models because we used the estimated emotions to calculate arousal and valence using equation \ref{eq:arousal}. Any error in emotion estimation can lead to a larger error in the circumplex model approach. Although the error is high for the circumplex model approach, it is important to look at how it is performing on trial data. 

\subsection{UnComfortability estimation for a single trial}
\begin{figure}[h!]
\centering
\begin{subfigure}{\linewidth}
	\includegraphics[width=\linewidth,keepaspectratio]{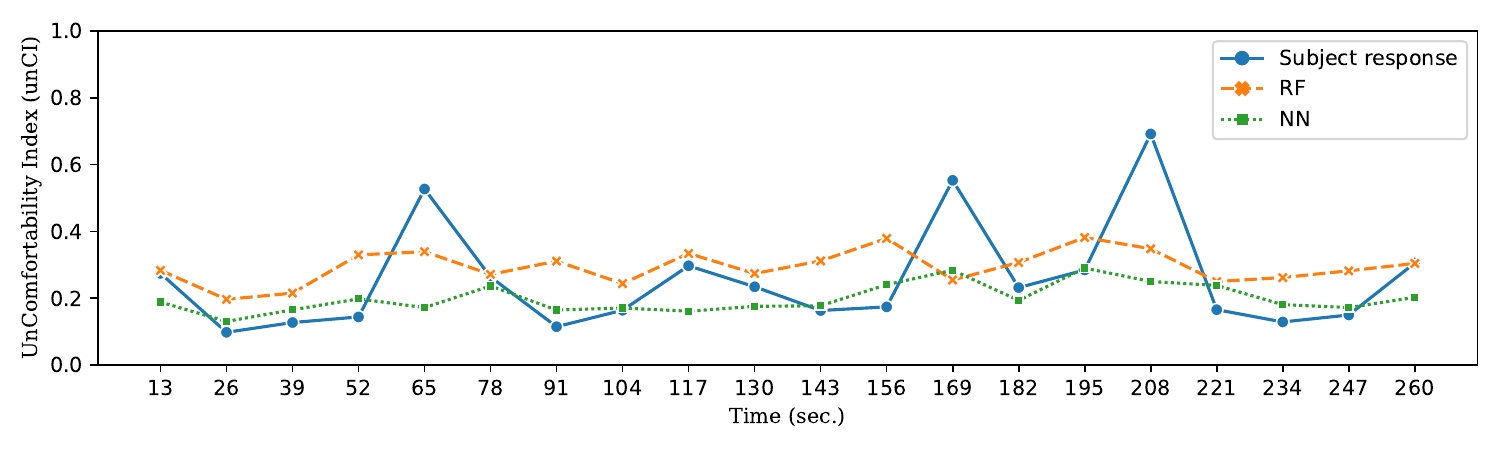}
	\caption{Performance of simple RF and NN that estimate Uncomfortability directly from physiological signals}
	\label{fig:one_trial_uncomfort_rf_nn}
\end{subfigure}
\begin{subfigure}{\linewidth}
	\includegraphics[width=\linewidth,keepaspectratio]{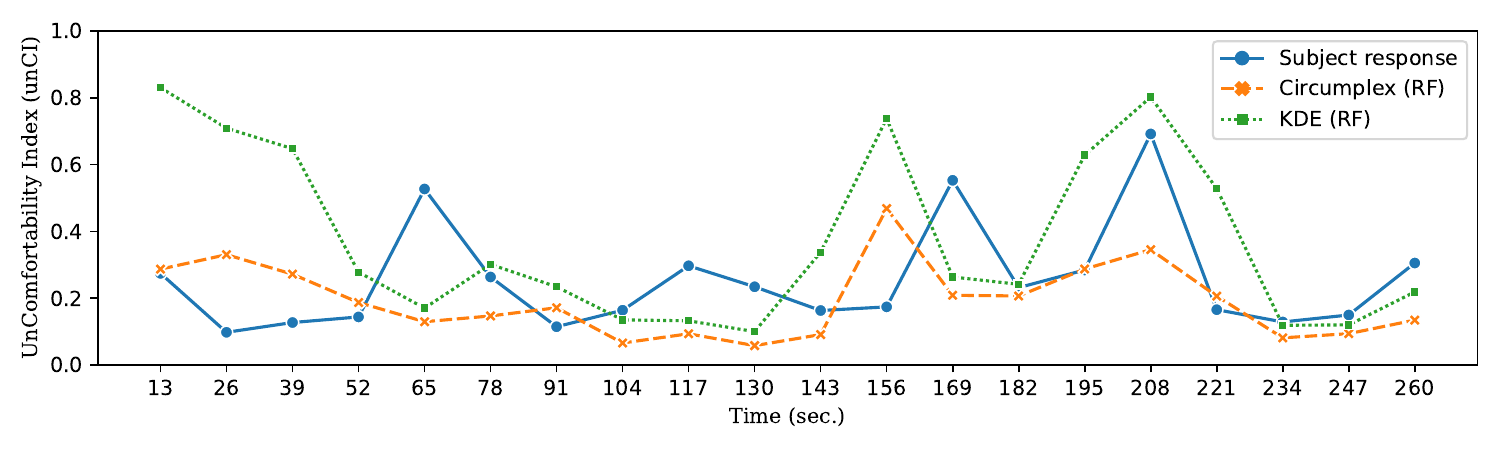}
	\caption{Performance of only Circumplex and KDE models that uses estimated emotions from RF regressors. }
	\label{fig:one_trial_uncomfort_circum_kde}
\end{subfigure}
\caption{Performance of both simple, circumplex, and KDE models estimation of UnComfortability Index for a trial }
\label{fig:figures}
\end{figure}

In this test, we apply the leave-one-out approach. We kept a trial and did not use it in the training of the models. The x-axis shows the time elapsed in a trial, and the y-axis shows the unCI level. During the trial, the robot's sensitivity was set to \textit{normal}, the trajectory was \textit{extreme}, and the velocity was \textit{fast}. The ground truth (reported by subjects) is shown in blue. The blue line shows that the subject uncomfort level changed during the trial. The data from a single trial fed into the ML models in order to estimate emotions, comfortability, and uncomfortability. Fig. \ref{fig:one_trial_uncomfort_rf_nn} shows the uncomfortability estimation from the two models (RF and NN). From Fig. \ref{fig:one_trial_uncomfort_rf_nn}, the simple approach that estimates unCI from physiological signal was not able to estimate unCI when it is high in three locations. The RF regressor is slightly better than the NN model.

On the other hand, the circumplex and KDE approaches shown in Fig. \ref{fig:one_trial_uncomfort_circum_kde} are better when estimating unCI. Even though the circumplex model has lower values, it captures the overall shape of the prediction. The KDE approach made a high unCI estimation where it was supposed to be low. This occurred due to limited data, where the estimation of emotion is done only using approximately 20 seconds of data. The model improved estimation once there was enough signal estimation. Although these two approaches had higher RMSE and MAE values, their estimation is better when estimating the high uncomfortability values. 

The proposed circumplex and KDE models use estimated emotion from the ML models to estimate \textit{arousal} and \textit{valence} and then estimate CI and unCI from them. However, it would be better if we estimated arousal and valence from the physiological signals. Since there is no standard way of calculating these metrics, and we did not ask participants about their arousal and valence levels, we cannot estimate directly. 

Although we estimated arousal and valence from emotion, this presents a huge opportunity as \textit{arousal} and \textit{valence} can be measured by physiological signals sensors and efficiently mapped to comfortability used in the circumplex model we created through these experiments.

\subsection{Real-Time estimation of unCI}
\label{sec:real-time-estimation}
\begin{figure}[h!]
	\centering
	\includegraphics[width=\linewidth,keepaspectratio]{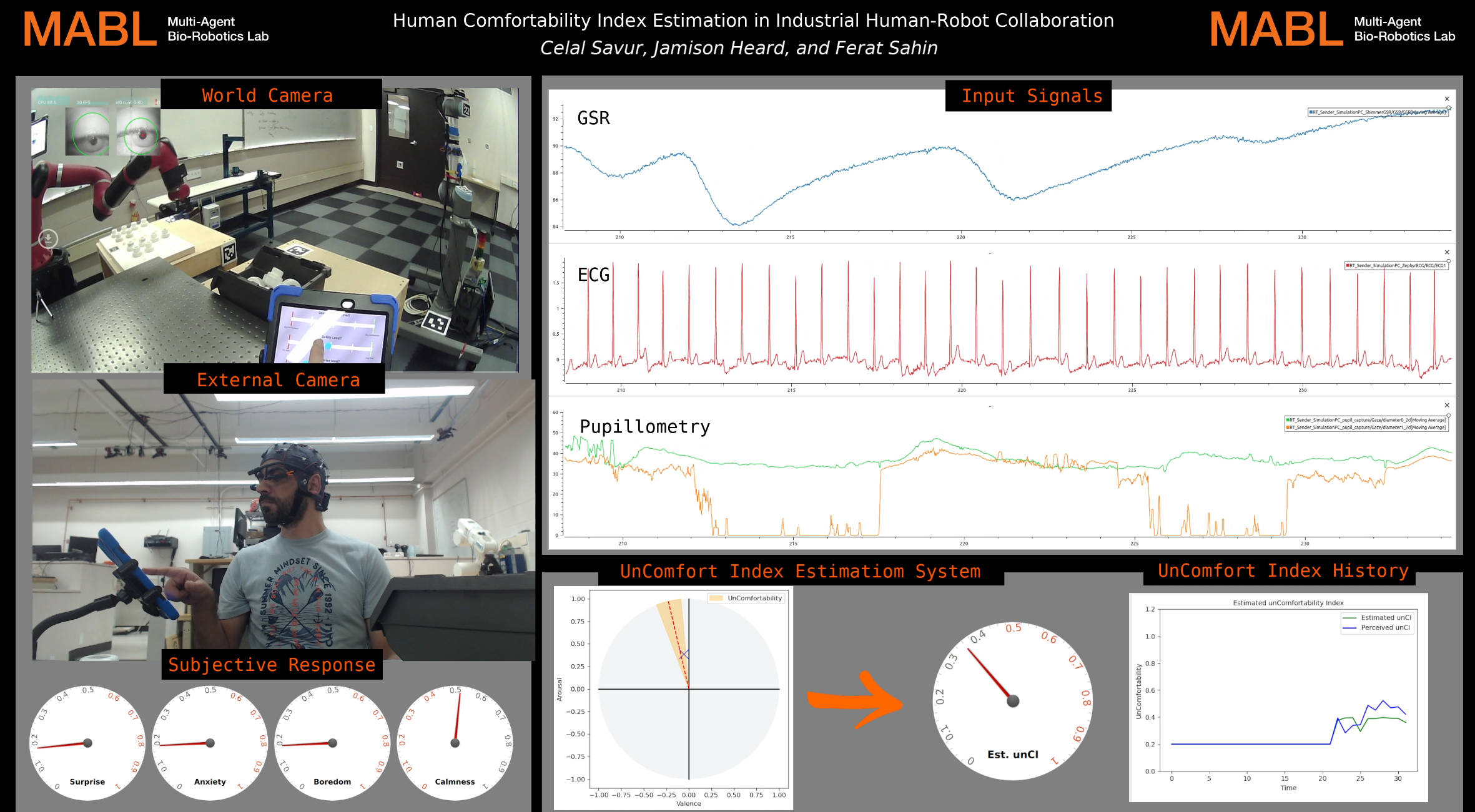}
	\caption{The world camera in the top-left corner shows the participant's view. External camera shows the participant. The gauges in the bottom corner shows the subjective responses are reported during trial. Inputs signals are the signal used in CIES system for real-time unCI estimation from GSR, ECG, and Pupil dilation signals by using circumplex model with RF regressor. The in the bottom-right corner shows the history of the estimated unCI with perceived unCI reported by the participant. The video for the trial can be access from ``https://youtu.be/qiTmN1ICVJo". }
	\label{fig:real-time-estimation}
\end{figure}

Controlling a robot during an experiment requires real-time estimation of unCI. Hence, an unCI estimation system (CIES) that estimates unCI from ECG, GSR, and pupil dilation signals in real time was developed. The system uses a ring buffer that adds new incoming signals to the buffer and removes older ones as soon as the buffer is full. The estimation of the unCI does not start until the buffer is full. Once the buffer is full, every second, the CIES make a copy of the buffer, clean, preprocess, and extract features from each signal separately. The extracted features are concatenated, then fed to the RF regressor model for emotion estimation. To have a better estimation, the last ten predictions were averaged. Thus, this allows the CIES to have smoother CI estimation. Fig. \ref{fig:real-time-estimation} shows the data streams of GSR, ECG, Pupil dilation, and the real-time estimation of unCI and perceived unCI that was reported from the subject. As shown in Fig. \ref{fig:real-time-estimation} history plot, the circumplex model with RF regressor estimates the perceived unCI with a small error. However, in order to validate the usefulness of the unCI estimation system, a new experiment is required where the participant will evaluate robots with CIES systems.

\section{Limitations}
\label{sec:limitations}
While the results show great promise, it should be noted that the study was conducted in a lab, a rather controlled environment. Subjects were essentially sedentary, and they only used one hand for the most part. Because it is generally recognized that motion artifacts affect physiological signals, data gathering in non-stationary conditions can be difficult. Future studies should consider motion artifacts and create algorithms to eliminate them.

The proposed physiological computing system used multiple sensors for physiological signals. This is necessary for the system to work; however, sometimes it may be difficult to use such sensors. For example, the proposed system using pupillometry signals requires the participant to wear special glasses. However, if the participant already uses glasses, then it becomes challenging to collect the signals. However, as the technology advances in the field of wearable sensors, this kind of challenge should be easier to deal with. 

In this research, arousal and valence were estimated from emotions. Because emotion estimation has noise, it is prone to error. A better approach would be to ask the subject about their arousal and valence using a self-assessment manikin (SAM) approach \cite{Bradley1994c, Koelstra2012}.  

Additionally, the human's comfort level could also be contingent upon their prior experience and familiarity with robots. These varying comfort levels could lead to suboptimal outcomes, including negative perceptions, reduced satisfaction, and poorer task results from HRC. Any positive or negative experience with robots can impact human comfort level, and this matters because human comfort level is tied to outcomes in terms of perceptions, experiences, and results.

\section{Conclusion}
\label{sec:conclusion}
In this article, we proposed two novel approaches for detecting and estimating human comfortability and uncomfortability using subjective responses and physiological signals. One of the proposed approaches was inspired by Russel's circumplex model, which allows emotion to be represented in terms of arousal and valence \cite{Russell1980}. The second approach was the estimation of CI/unCI from arousal and valence by fitting a KDE distribution. To estimate arousal and valence, emotions were estimated from physiological signals. The results of this study may be useful for future systems. Leveraging estimations of human comfort can be utilized to personalize robot behavior and safety-response systems to achieve more optimal human-robot collaboration experiences and outcomes.

In the next experiment, we will use the trained ML model in this research to estimate uncomfortability. The estimated uncomfortability will be used to modify the robot's velocity and cushioning distance between the human and the robot. As the human becomes less comfortable the robot will reduce its velocity and increase the cushioning distance and the robot will go to its default settings as the human becomes comfortable. 

The dataset for the experiment will be available for other researchers to test and develop their ML models for CI or unCI estimation or other application. The dataset will be available at https://mabl.rit.edu/dataset. 

\section*{Acknowledgments}
The authors would like to thank Abhiraj Patil for his help during data collection. This material is based upon work partially supported by the National Science Foundation under Award No. DGE-2125362. Any opinions, findings, and conclusions or recommendations expressed in this material are those of the author(s) and do not necessarily reflect the views of the National Science Foundation.

\ifCLASSOPTIONcaptionsoff
  \newpage
\fi



%

\bibliographystyle{IEEEtran}
\bibliography{survey}

\begin{thebibliography}{10}
\providecommand{\url}[1]{#1}
\csname url@samestyle\endcsname
\providecommand{\newblock}{\relax}
\providecommand{\bibinfo}[2]{#2}
\providecommand{\BIBentrySTDinterwordspacing}{\spaceskip=0pt\relax}
\providecommand{\BIBentryALTinterwordstretchfactor}{4}
\providecommand{\BIBentryALTinterwordspacing}{\spaceskip=\fontdimen2\font plus
\BIBentryALTinterwordstretchfactor\fontdimen3\font minus
  \fontdimen4\font\relax}
\providecommand{\BIBforeignlanguage}[2]{{%
\expandafter\ifx\csname l@#1\endcsname\relax
\typeout{** WARNING: IEEEtran.bst: No hyphenation pattern has been}%
\typeout{** loaded for the language `#1'. Using the pattern for}%
\typeout{** the default language instead.}%
\else
\language=\csname l@#1\endcsname
\fi
#2}}
\providecommand{\BIBdecl}{\relax}
\BIBdecl

\bibitem{Absolutereports2019}
\BIBentryALTinterwordspacing
Absolutereports, ``{Global Collaborative Robots Market 2019 by Manufacturers,
  Regions, Type and Application, Forecast to 2024},'' Tech. Rep., 2019.
  [Online]. Available:
  \url{https://www.marketwatch.com/press-release/collaborative-robots-market-size-2022-research-report-including-industry-analysis-geographical-regions-manufactures-opportunities-growth-factors-and-forecast-to-2024-says-absolute-reports-2022-01-19}
\BIBentrySTDinterwordspacing

\bibitem{ISO2016}
\BIBentryALTinterwordspacing
{ISO 2016}, ``{ISO/TS 15066:2016},'' 2016. [Online]. Available:
  \url{https://tinyurl.com/ebfesr2j}
\BIBentrySTDinterwordspacing

\bibitem{Kumar2021}
S.~Kumar, C.~Savur, and F.~Sahin, ``{Survey of Human–Robot Collaboration in
  Industrial Settings: Awareness, Intelligence, and Compliance},'' \emph{IEEE
  Transactions on Systems, Man, and Cybernetics: Systems}, vol.~51, no.~1, pp.
  280--297, jan 2021.

\bibitem{NSF2019}
\BIBentryALTinterwordspacing
NSF, ``{Information and Intelligent Systems (IIS): Core Programs},'' 2019.
  [Online]. Available:
  \url{https://www.nsf.gov/pubs/2018/nsf18570/nsf18570.htm}
\BIBentrySTDinterwordspacing

\bibitem{Fairclough2017}
\BIBentryALTinterwordspacing
S.~H. Fairclough, ``{Physiological Computing and Intelligent Adaptation},'' in
  \emph{Emotions and Affect in Human Factors and Human-Computer
  Interaction}.\hskip 1em plus 0.5em minus 0.4em\relax Elsevier, 2017, no.
  2017, pp. 539--556. [Online]. Available:
  \url{https://linkinghub.elsevier.com/retrieve/pii/B9780128018514000203}
\BIBentrySTDinterwordspacing

\bibitem{Kulic2007}
D.~Kulic and E.~A. Croft, ``{Affective State Estimation for Human–Robot
  Interaction},'' \emph{IEEE Transactions on Robotics}, vol.~23, no.~5, pp.
  991--1000, oct 2007.

\bibitem{Kulic2007a}
D.~Kuli{\'{c}} and E.~Croft, ``{Physiological and subjective responses to
  articulated robot motion},'' \emph{Robotica}, vol.~25, no.~1, pp. 13--27, jan
  2007.

\bibitem{Hu2016}
\BIBentryALTinterwordspacing
W.~L. Hu, K.~Akash, N.~Jain, and T.~Reid, ``{Real-Time Sensing of Trust in
  Human-Machine Interactions},'' \emph{IFAC-PapersOnLine}, vol.~49, no.~32, pp.
  48--53, 2016. [Online]. Available:
  \url{http://dx.doi.org/10.1016/j.ifacol.2016.12.188}
\BIBentrySTDinterwordspacing

\bibitem{Kulic2005}
D.~Kulic and E.~A. Croft, ``{Anxiety detection during human-robot
  interaction},'' in \emph{2005 IEEE/RSJ International Conference on
  Intelligent Robots and Systems}.\hskip 1em plus 0.5em minus 0.4em\relax IEEE,
  2005, pp. 616--621.

\bibitem{Kulic2005a}
------, ``{Real-time safety for human - robot interaction},'' in \emph{ICAR
  '05. Proceedings., 12th International Conference on Advanced Robotics,
  2005.}, vol. 2005.\hskip 1em plus 0.5em minus 0.4em\relax IEEE, 2005, pp.
  719--724.

\bibitem{Kulic2006}
D.~Kulic and E.~Croft, ``{Estimating Robot Induced Affective State using Hidden
  Markov Models},'' in \emph{ROMAN 2006 - The 15th IEEE International Symposium
  on Robot and Human Interactive Communication}.\hskip 1em plus 0.5em minus
  0.4em\relax IEEE, sep 2006, pp. 257--262.

\bibitem{Rani2007}
P.~Rani, N.~Sarkar, and J.~Adams, ``{Anxiety-based affective communication for
  implicit human–machine interaction},'' \emph{Advanced Engineering
  Informatics}, vol.~21, no.~3, pp. 323--334, jul 2007.

\bibitem{Villani2018a}
\BIBentryALTinterwordspacing
V.~Villani, L.~Sabattini, C.~Secchi, and C.~Fantuzzi, ``{A Framework for
  Affect-Based Natural Human-Robot Interaction},'' in \emph{2018 27th IEEE
  International Symposium on Robot and Human Interactive Communication
  (RO-MAN)}.\hskip 1em plus 0.5em minus 0.4em\relax IEEE, aug 2018, pp.
  1038--1044. [Online]. Available:
  \url{https://ieeexplore.ieee.org/document/8525658/}
\BIBentrySTDinterwordspacing

\bibitem{Dobbins2018}
\BIBentryALTinterwordspacing
C.~Dobbins, S.~Fairclough, P.~Lisboa, and F.~F.~G. Navarro, ``{A Lifelogging
  Platform Towards Detecting Negative Emotions in Everyday Life using Wearable
  Devices},'' in \emph{2018 IEEE International Conference on Pervasive
  Computing and Communications Workshops (PerCom Workshops)}.\hskip 1em plus
  0.5em minus 0.4em\relax IEEE, mar 2018, pp. 306--311. [Online]. Available:
  \url{https://ieeexplore.ieee.org/document/8480180/}
\BIBentrySTDinterwordspacing

\bibitem{Too2009}
J.~Too, C.~Tan, F.~Duan, Y.~Zhang, K.~Watanabe, R.~Kato, and T.~Arai,
  ``{Human-Robot Collaboration in Cellular Manufacturing: Design and
  Development},'' \emph{International Conference on Intelligent Robots and
  Systems}, pp. 29--34, 2009.

\bibitem{Heard2019}
J.~Heard and J.~A. Adams, ``{Multi-Dimensional Human Workload Assessment for
  Supervisory Human–Machine Teams},'' \emph{Journal of Cognitive Engineering
  and Decision Making}, vol.~13, no.~3, pp. 146--170, 2019.

\bibitem{Landi2018}
C.~T. Landi, V.~Villani, F.~Ferraguti, L.~Sabattini, C.~Secchi, and
  C.~Fantuzzi, ``{Relieving operators' workload: Towards affective robotics in
  industrial scenarios},'' \emph{Mechatronics}, vol.~54, no. April, pp.
  144--154, 2018.

\bibitem{Rani2006}
P.~Rani, N.~Sarkar, and C.~Liu, ``{Maintaining Optimal Challenge in Computer
  Games through Real-Time Physiological Feedback Mechanical Engineering},''
  \emph{Task-Specific Information Processing in Operational and Virtual
  Environments: Foundations of Augmented Cognition}, pp. 184--192, 2006.

\bibitem{Liu2006}
C.~Liu, P.~Rani, and N.~Sarkar, ``{Human-robot interaction using affective
  cues},'' \emph{Proceedings - IEEE International Workshop on Robot and Human
  Interactive Communication}, pp. 285--290, 2006.

\bibitem{Amin2019a}
\BIBentryALTinterwordspacing
M.~R. Amin and R.~T. Faghih, ``{Robust Inference of Autonomic Nervous System
  Activation Using Skin Conductance Measurements: A Multi-Channel Sparse System
  Identification Approach},'' \emph{IEEE Access}, vol.~7, pp.
  173\,419--173\,437, 2019. [Online]. Available:
  \url{https://ieeexplore.ieee.org/document/8917550/}
\BIBentrySTDinterwordspacing

\bibitem{Wickramasuriya2018}
D.~S. Wickramasuriya, C.~Qi, and R.~T. Faghih, ``{A State-Space Approach for
  Detecting Stress from Electrodermal Activity},'' \emph{Proceedings of the
  Annual International Conference of the IEEE Engineering in Medicine and
  Biology Society, EMBS}, vol. 2018-July, pp. 3562--3567, 2018.

\bibitem{Wickramasuriya2020}
D.~S. Wickramasuriya and R.~T. Faghih, ``{A Marked Point Process Filtering
  Approach for Tracking Sympathetic Arousal from Skin Conductance},''
  \emph{IEEE Access}, vol.~8, pp. 68\,499--68\,513, 2020.

\bibitem{Russell1980}
J.~A. Russell, ``{A circumplex model of affect},'' \emph{Journal of Personality
  and Social Psychology}, vol.~39, no.~6, pp. 1161--1178, 1980.

\bibitem{Toisoul2021}
A.~Toisoul, J.~Kossaifi, A.~Bulat, G.~Tzimiropoulos, and M.~Pantic,
  ``{Estimation of continuous valence and arousal levels from faces in
  naturalistic conditions},'' \emph{Nature Machine Intelligence}, vol.~3,
  no.~1, pp. 42--50, jan 2021.

\bibitem{Mollahosseini2017}
A.~Mollahosseini, B.~Hasani, and M.~H. Mahoor, ``{AffectNet: A database for
  facial expression, valence, and arousal computing in the wild},''
  \emph{arXiv}, vol.~10, no.~1, pp. 18--31, 2017.

\bibitem{Elena2020}
M.~Elena, L.~Redondo, A.~Vignolo, R.~Niewiadomski, F.~Rea, and A.~Sciutti,
  \emph{{Comfortability Levels ?}}\hskip 1em plus 0.5em minus 0.4em\relax
  Springer International Publishing, 2020, vol.~2.

\bibitem{Du2020}
N.~Du, F.~Zhou, E.~M. Pulver, D.~M. Tilbury, L.~P. Robert, A.~K. Pradhan, and
  X.~J. Yang, ``{Examining the effects of emotional valence and arousal on
  takeover performance in conditionally automated driving},''
  \emph{Transportation Research Part C: Emerging Technologies}, vol. 112, no.
  September 2019, pp. 78--87, 2020.

\bibitem{Pedregosa2011}
F.~Pedregosa, G.~Varoquaux, A.~Gramfort, V.~Michel, B.~Thirion, O.~Grisel,
  M.~Blondel, A.~M{\"{u}}ller, J.~Nothman, G.~Louppe, P.~Prettenhofer,
  R.~Weiss, V.~Dubourg, J.~Vanderplas, A.~Passos, D.~Cournapeau, M.~Brucher,
  M.~Perrot, and {\'{E}}.~Duchesnay, ``{Scikit-learn: Machine Learning in
  Python},'' \emph{Journal of Machine Learning Research}, vol.~12, pp.
  2825--2830, jan 2012.

\bibitem{Villani2018}
\BIBentryALTinterwordspacing
V.~Villani, F.~Pini, F.~Leali, and C.~Secchi, ``{Survey on human–robot
  collaboration in industrial settings: Safety, intuitive interfaces and
  applications},'' \emph{Mechatronics}, vol.~55, no. March, pp. 248--266, nov
  2018. [Online]. Available:
  \url{https://doi.org/10.1016/j.mechatronics.2018.02.009
  https://linkinghub.elsevier.com/retrieve/pii/S0957415818300321}
\BIBentrySTDinterwordspacing

\bibitem{Boucsein2012}
W.~Boucsein, \emph{{Electrodermal Activity}}.\hskip 1em plus 0.5em minus
  0.4em\relax Boston, MA: Springer US, 1992.

\bibitem{Bradley1994a}
M.~M. Bradley and P.~J. Lang, ``{Measuring emotion: The self-assessment manikin
  and the semantic differential},'' \emph{Journal of Behavior Therapy and
  Experimental Psychiatry}, vol.~25, no.~1, pp. 49--59, 1994.

\bibitem{Yik2011}
M.~Yik, J.~A. Russell, and J.~H. Steiger, ``{A 12-Point Circumplex Structure of
  Core Affect},'' \emph{Emotion}, vol.~11, no.~4, pp. 705--731, 2011.

\bibitem{Carreiras}
\BIBentryALTinterwordspacing
C.~Carreiras, A.~P. Alves, A.~Louren\c{c}o, F.~Canento, H.~Silva, and A.~Fred,
  ``{BioSPPy: Biosignal Processing in Python}.'' [Online]. Available:
  \url{https://github.com/PIA-Group/BioSPPy/}
\BIBentrySTDinterwordspacing

\bibitem{Gamboa2008}
H.~F.~S. Gamboa, ``{Multi-modal behavioral biometrics based on HCI and
  electrophysiology},'' Ph.D. dissertation, UNIVERSIDADE TECNICA DE LISBOA,
  2008.

\bibitem{Bonifacci2015}
P.~Bonifacci, L.~Desideri, and C.~Ottaviani, ``{Familiarity of Faces: Sense or
  Feeling?}'' \emph{Journal of Psychophysiology}, vol.~29, no.~1, pp. 20--25,
  jan 2015.

\bibitem{Mathot2018}
S.~Math{\^{o}}t, ``{Pupillometry: Psychology, Physiology, and Function},''
  \emph{Journal of Cognition}, vol.~1, no.~1, pp. 1--23, feb 2018.

\bibitem{Paprocki2017}
R.~Paprocki and A.~Lenskiy, ``{What Does Eye-Blink Rate Variability Dynamics
  Tell Us About Cognitive Performance?}'' \emph{Frontiers in Human
  Neuroscience}, vol.~11, no. December, pp. 1--9, dec 2017.

\bibitem{Ethem2014}
A.~Ethem, \emph{{Introduction to Machine Learning}}, 3rd~ed.\hskip 1em plus
  0.5em minus 0.4em\relax MIT, 2014.

\bibitem{BAUER2008}
A.~BAUER, D.~WOLLHERR, and M.~BUSS, ``{HUMAN–ROBOT COLLABORATION: A
  SURVEY},'' \emph{International Journal of Humanoid Robotics}, vol.~05,
  no.~01, pp. 47--66, mar 2008.

\bibitem{Sahin2022}
M.~Sahin and C.~Savur, ``{Evaluation of Human Perceived Safety during HRC Task
  using Multiple Data Collection Methods},'' in \emph{2022 17th Annual
  Conference System of Systems Engineering, SoSE 2022}, 2022, pp. 465--470.

\bibitem{Mejia-Mejia2020}
E.~Mej{\'{i}}a-Mej{\'{i}}a, J.~M. May, R.~Torres, and P.~A. Kyriacou, ``{Pulse
  rate variability in cardiovascular health: a review on its applications and
  relationship with heart rate variability},'' \emph{Physiological
  Measurement}, vol.~41, no.~7, p. 07TR01, aug 2020.

\bibitem{Kassner2014}
M.~Kassner, W.~Patera, and A.~Bulling, ``{Pupil: An Open Source Platform for
  Pervasive Eye Tracking and Mobile Gaze-based Interaction},'' in
  \emph{Proceedings of the 2014 ACM International Joint Conference on Pervasive
  and Ubiquitous Computing: Adjunct Publication}.\hskip 1em plus 0.5em minus
  0.4em\relax New York, NY, USA: ACM, sep 2014, pp. 1151--1160.

\bibitem{Savur2019}
C.~Savur, S.~Kumar, S.~Arora, T.~Hazbar, and F.~Sahin, ``{HRC-SoS: Human robot
  collaboration experimentation platform as system of systems},'' \emph{2019
  14th Annual Conference System of Systems Engineering, SoSE 2019}, pp.
  206--211, 2019.

\bibitem{SCCN2018}
SCCN, ``{Lab Stream Layer (LSL)},'' 2018.

\bibitem{Savur2019a}
C.~Savur, S.~Kumar, and F.~Sahin, ``{A framework for monitoring human
  physiological response during human robot collaborative task},''
  \emph{Conference Proceedings - IEEE International Conference on Systems, Man
  and Cybernetics}, vol. 2019-Octob, pp. 385--390, 2019.

\bibitem{Bradley1994c}
M.~M. Bradley and P.~J. Lang, ``{Measuring emotion: The self-assessment manikin
  and the semantic differential},'' \emph{Journal of Behavior Therapy and
  Experimental Psychiatry}, vol.~25, no.~1, pp. 49--59, mar 1994.

\bibitem{Koelstra2012}
S.~Koelstra, C.~M{\"{u}}hl, M.~Soleymani, J.~S. Lee, A.~Yazdani, T.~Ebrahimi,
  T.~Pun, A.~Nijholt, and I.~Patras, ``{DEAP: A database for emotion analysis;
  Using physiological signals},'' \emph{IEEE Transactions on Affective
  Computing}, vol.~3, no.~1, pp. 18--31, 2012.

\end{thebibliography}

\end{document}